\pdfoutput=1
\documentclass[11pt]{article}

\usepackage{EACL2023}

\usepackage{times}
\usepackage{latexsym}
\usepackage{graphicx}
\usepackage[T1]{fontenc}
\usepackage[utf8]{inputenc}

\usepackage{inconsolata}
\usepackage{microtype}
\usepackage{enumitem}
\usepackage{amsmath}
\usepackage{amssymb}
\usepackage{multirow}
\usepackage{multicol}
\usepackage{enumitem}
\usepackage{xcolor}
\usepackage{bm}
\usepackage{booktabs}
\usepackage{url}
\title{PANACEA: An Automated Misinformation Detection System on COVID-19}

\author{ Runcong Zhao$^{1,3}$, Miguel Arana-Catania$^5$,  Lixing Zhu$^{1,3}$, Elena Kochkina$^{2,4}$, \\ 
\textbf{Lin Gui$^3$, Arkaitz Zubiaga$^2$, Rob Procter$^{1,4}$, Maria Liakata$^{1,2,4}$, Yulan He$^{1,3,4}$}\\
  $^1$University of Warwick, $^2$Queen Mary University of London\\
  $^3$King's College London,  $^4$The Alan Turing Institute,  $^5$Cranfield University\\
  \texttt{\{runcong.zhao, yulan.he\}@kcl.ac.uk} }

\begin{document}
\maketitle
\begin{abstract}

%The misinformation that always follows the spread of information on various media continues to grow and harm the public. 
In this demo, we introduce a web-based misinformation detection system PANACEA on COVID-19 related claims, which has two modules, \emph{fact-checking} and \emph{rumour detection}. %, corresponding to two types of misinformation, \emph{false} and \emph{intents to harm}. 
%At present, misinformation detection sites 
%such as EVIDENCEMINER, PubMed, BBC Reality Check, and Full Fact,
%are designed to retrieve related results from reliable knowledge bases, which may not be available at the early stages of fact checking rumours, for example. %In addition, misinformation with \emph{intents to harm}, such as misinterpretation, quote out of context, are often ignored by current systems as it might be aligned with reliable knowledge bases. 
Our \emph{fact-checking} module, which is supported by novel natural language inference methods with a self-attention network, outperforms state-of-the-art approaches. It is also able to give automated veracity assessment and ranked supporting evidence with the stance towards the claim to be checked. In addition, PANACEA adapts the bi-directional graph convolutional networks model, %which is based on comment networks of related tweets, to detect rumours without relying on the knowledge base. 
which is able to detect rumours based on comment networks of related tweets, instead of relying on the knowledge base. 
%This \emph{rumour detection} module assists to warn users of misinformation missed by \emph{fact checking}.
This \emph{rumour detection} module assists by warning the users in the early stages when a knowledge base may not be available.

\end{abstract}

\section{Introduction}
%People are alerted to be more sensitised to the danger of misinformation during the COVID-19 pandemic, as propagation of misinformation has become prevalent \cite{misinfo2020enders, misinfo2020kouzy}.
The dangers of misinformation have become even more apparent to the general public during the COVID-19 pandemic. Following \emph{false} treatment information has led to a high number of deaths and hospitalisations~\cite{islam2020covid}. Manual verification can not  scale to the amount of misinformation being spread, therefore there is a need to develop automated tools to assist in this process. 

In this work, we focus on automating  misinformation detection using information from credible sources as well as social media. We produce a web-based tool that can be used by the general public to inspect relevant information about the claims that they want to check, see supporting or refuting evidence, and social media propagation patterns. %: one being \emph{false} and another one being \emph{intent to harm}.
%Misinformation can have very harmful consequences. For instance, \emph{false} treatment information such as ``chloroquine can cure coronavirus'' can put public safety in danger and increase pressure on the health system.  %and for \emph{intent to harm}, partially showing information such as ``vaccine causes death'' without mentioning the very low probability, could obstruct people’s willingness to get vaccinated and increase the chance of spreading the disease. Those types of misinformation can strongly influence human behaviour and negatively impact public health interventions, thus it is vital to detect misinformation in a timely manner. 
%and other misinformation, %for \emph{intent to harm}, 
%partially showing information such as ``vaccine causes death'' without mentioning that the benefits outweigh the risks, could obstruct people’s willingness to get vaccinated and increase the chance of spreading the disease. 
%Misinformation can strongly influence human behaviour and negatively impact public health interventions, thus it is vital to detect misinformation in a timely manner. 

%One promising approach for tackling this problem is to develop tools for automated misinformation detection. 
For \emph{false} information, the commonly used and relatively reliable method for automated veracity assessment is to check the claim against a verified knowledge base, which we call \emph{fact-checking}. Previous works such as EVIDENCEMINER \cite{EVIDENCEMINER2020}, PubMed\footnote{\url{https://www.ncbi.nlm.nih.gov/pmc/}}  %\url{https://pubmed.ncbi.nlm.nih.gov/}} 
and COVID-19 fact-checking sites recommended by the NHS\footnote{\url{https://library.hee.nhs.uk/covid-19/coronavirus-\%28covid-19\%29-misinformation}} are all designed to retrieve related documents/sentences from a reliable knowledge base. 
However, %no veracity assessment 
this approach leaves users to summarise a large amount of potentially conflicting evidence themselves.  % instead of considering the whole query, therefore the result can only be ranked by their relevance. 
PANACEA, which is supported by novel natural language inference methods \cite{nli2022}, is instead able to provide automated veracity assessment and supporting evidence for the input claim. In addition, previous works retrieve results using entities in the input claim, and thus often include results related to a keyword in the input claim instead of the whole query, while PANACEA considers the whole query for better result. The supporting pieces of evidence are also ranked by their relevance score and classified according to their stance towards the input claim. 
%Results are calculated at the backend within minutes with a queuing system designed to manage GPU resources for faster fact checking and running multiple requests. 

In addition to \emph{false} information, %\emph{true} information can also be weaponized using it with an \emph{intent to harm} the public. 
truthful information can also be misused to harm competitors or gain attention on social media \cite{whymisinfor2020, whymisinfor2020tsfati}. However, the latter is harder to be found by checking reliable knowledge bases as those are focused on \emph{false} information. %To the best of our knowledge, there is no existing system that can reliably identify this type of misinformation. 
Regarding this issue, previous work has analysed the spread of misinformation using features such as  stance \cite{covidmisinfo2021}, sentiment, topics, geographical spread, the reliability of external links included in the tweet \cite{covidmisinfo2020}, origin and propagation networks \cite{twittertrails2014}. However, it is still hard for users to identify rumours by directly looking at those features.
%Efforts are put into visualising the statistics of misinformation about tweets such as the sentiments, topics, spread across countries \cite{covidmisinfo2020}, origin and propagation network \cite{twittertrails2014}, which aim to assist misinformation identification. 
 %Previous works have tried to identify misinformation tweets by checking the reliability of external links including in the tweet \cite{covidmisinfo2020}. Unfortunately, most tweets do not include such links and not all links are accessible for reliability checking. 
 Previous research shows that the propagation pattern is different between fake and real news, which  would offer additional features for early detection of misinformation on social media \cite{fakenews2020}. PANACEA extends this by using tweets’ propagation patterns to identify rumours.
 %PANACEA extends this research of \emph{Rumour detection} by using the tweets’ propagation pattern. 
 \emph{Rumour detection} is not as reliable as \emph{fact-checking}, but it generalises the system to various situations that \emph{fact-checking} cannot cover: First, \emph{true} or \emph{unverified} information with \emph{intent to harm}; Second, scenarios where no verified knowledge database is available. \emph{Rumour detection} cannot prove the truth of a claim but may alert the user about claims with a high risk of being misinformation.
 %Results are calculated at the backend by retrieving related tweets’ propagation patterns using BM25. 
 
 Previous work have either retrieved tweets from a short fixed time period \cite{covidmisinfo2020} or search recent tweets \cite{twittertrails2014}, which is limited by Twitter to only the last 7 days. We instead maintain an updated database which is constituted of an annotated tweets dataset with popular claims and an unlabelled streaming of COVID-19 related tweets that are crawled and selected periodically to update the dataset. Besides building on the various analytic functionalities used in previous work, PANACEA improves the architecture of these elements and adds extra features to the updated dataset for more efficient results. 
 
 %Our system is publicly available online \footnote{\url{https://panacea.dcs.warwick.ac.uk/search}}. 
 %Our system is publicly available online \footnote{\url{https://panacea2020.github.io/}}
 A screencast video introducing the system\footnote{\url{https://www.youtube.com/watch?v=D1PN8_9oYso}}, illustrating its use in the checking of a COVID-19 claim, and the demo%\footnote{\url{https://github.com/BLPXSPG/PANACEA}} 
\footnote{\url{https://panacea2020.github.io/}}
 are also available online. The system can be easily adapted to other claim topics.
 
 PANACEA covers various types of misinformation detection related to COVID-19 with the following contributions:
\begin{itemize}%[noitemsep]
    \item We built a new web-based system, PANACEA, which is able to perform both \emph{fact-checking} and \emph{rumour detection} with natural language claims submitted by users. %This covers broader types of misinformation compared with previous works. 
    The system includes visualisations of various statistical analyses of the results for a better user understanding.
    \item PANACEA performs automated veracity assessment and provides supporting evidence that can be ranked by various criteria, supported by novel natural language inference methods. The system is able to manage multiple user requests with low latency thanks to our development of a queuing system. 
    \item PANACEA is able to perform automated rumour detection by exploiting state-of-the-art research on propagation patterns. The system uses an annotated dataset and streams of COVID-19 tweets are collected to maintain an updated database. 
\end{itemize}

\section{Datasets}
The following datasets are used in the project:
\paragraph{Knowledge Database}
This is used for fact-checking, and includes COVID-19 related documents from selected reliable sources \footnote{\emph{Centers for Disease Control and Prevention} (CDC), \emph{European Centre for Disease Prevention and Control} (ECDC), \emph{WebMD} and \emph{World Health Organisation} (WHO)}. 
The documents were cleaned and split into 300 token paragraphs to construct a reliable knowledge database, whose supporting documents are retrieved and visualised in our system.
\paragraph{PANACEA Dataset} \cite{nli2022}, constructed from COVID-19 related data sources\footnote{Corona VirusFacts Database, CoAID dataset \cite{coaid2020}, MM-COVID \cite{mmcovid2020}, CovidLies \cite{covidlies2020}, TREC Health Misinformation track and TREC COVID challenge \cite{trec2021}} and using BM25 and MonoT5 \cite{monot5} to remove duplicate claims. This dataset includes 5,143 labelled claims (1,810 \emph{False} and 3,333 \emph{True}), and their respective text, source and claim sub-type. 
\paragraph{COVID-RV dataset} 
In order to fine-tune our model, we constructed a new COVID-19 related propagation tree dataset for rumour detection. Similar previous datasets are Twitter15 and Twitter16 \cite{tdrvnn2018}, which are widespread tweets’ propagation trees with rumour labels, however, they are not COVID-19 related. Our dataset has been constructed by extending COVID-RV \cite{rumourdetection2022}, including %Retweet are excluded because it does not have any extra information, but \emph{the number of retweet}, is recorded along with 
\emph{the number of retweets}, \emph{user id}, \emph{post time}, \emph{text}, \emph{location} and \emph{tweet reply ids} as metadata for each tweet. Each tree is annotated with a related claim chosen from our claim dataset and a stance label (chosen from \emph{Support} or \emph{Refute}) towards its related claim. Such a stance label for each tree is purely based on the content of the source tweet. In COVID-RV the conversations are annotated as either \emph{True} or \emph{False} based on the veracity of the claim and the stance of the source tweet towards it. Tweets supporting a false claim or challenging a true claim are annotated as \emph{False}, tweets supporting true claims or challenging a false claim are annotated as \emph{True}. Twitter15 and Twitter16 datasets also contain \emph{Unverified} conversations, which are discussing claim that are neither confirmed or denied. 
%711 widespread source tweets related to popular claims

\paragraph{COVID Twitter Propagation Tree (Live)} 
Besides the last dataset constructed for fine-tuning, PANACEA also runs a crawler to collect a stream of COVID-19 tweets that are used to maintain an updated database. This live dataset is not annotated, instead, it is labelled by the pre-trained rumour detection model. As the Twitter’s search API does not allow retrieval of tweets beyond a week window, we retrieve COVID-19 related historical tweets based on the widely used dataset of COVID-19-TweetIDs \cite{tweetid}, which contains more than 1 billion tweet IDs. Considering the size of the dataset, and for the storage and retrieval efficiency, we filtered out the less popular tweets with limited impact. To date, more than 12k propagation trees have been collected, starting from January 2020. %Metadata collected for each tweet is the same as the annotated tweet and 
For each tweet, its pseudo rumour label is generated by the trained model.

\section{Architecture of PANACEA}
\begin{figure}[h]
    \centering
    \includegraphics[width=1\linewidth]{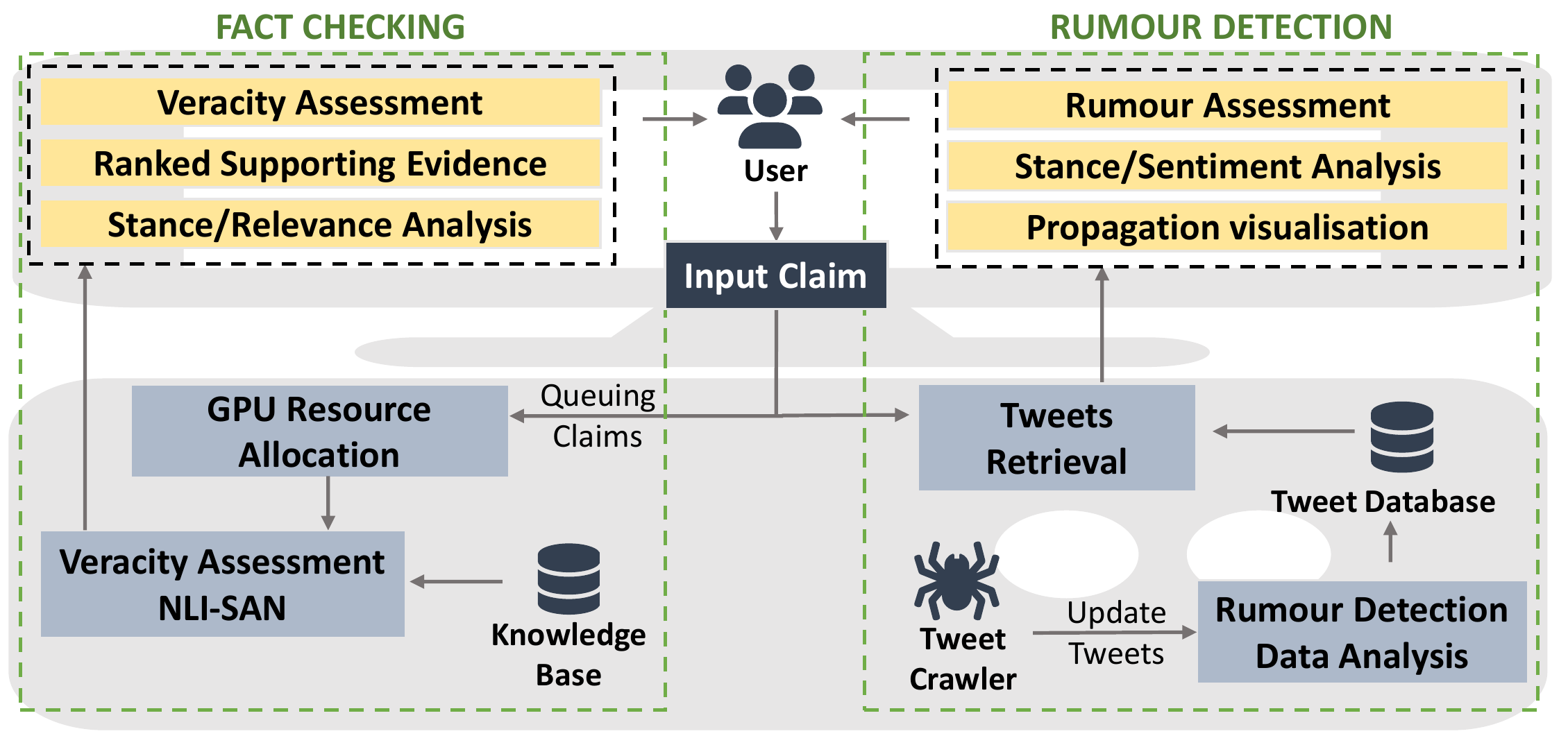}
    \caption{Architecture of PANACEA}
    \label{fig:architecture}
\end{figure}

Figure~\ref{fig:architecture} shows an overview of PANACEA, including two functions: \emph{fact-checking} and \emph{rumour detection} for COVID-19. For \emph{fact-checking}, there are three modules: (1) resource allocation system; (2) veracity assessment; and (3) supporting evidence retrieval. PANACEA also supports a unique function, \emph{rumour detection} by propagation patterns, which has the following modules: (1) tweet retrieval; (2) rumour detection; and (3) tweet meta-information analysis. %The Code is available at \url{https://github.com/BLPXSPG/PANACEA}. 

\subsection{Fact-Checking}

%\begin{figure}[htb]
%    \centering
%    \includegraphics[width=0.8\linewidth]{autocomplete.png}
%    \caption{Autocompletion function of input claims}
%    \label{fig:autocomplete}
%\end{figure}

\begin{figure*}[htb]
    \centering
    \includegraphics[width=0.95\linewidth]{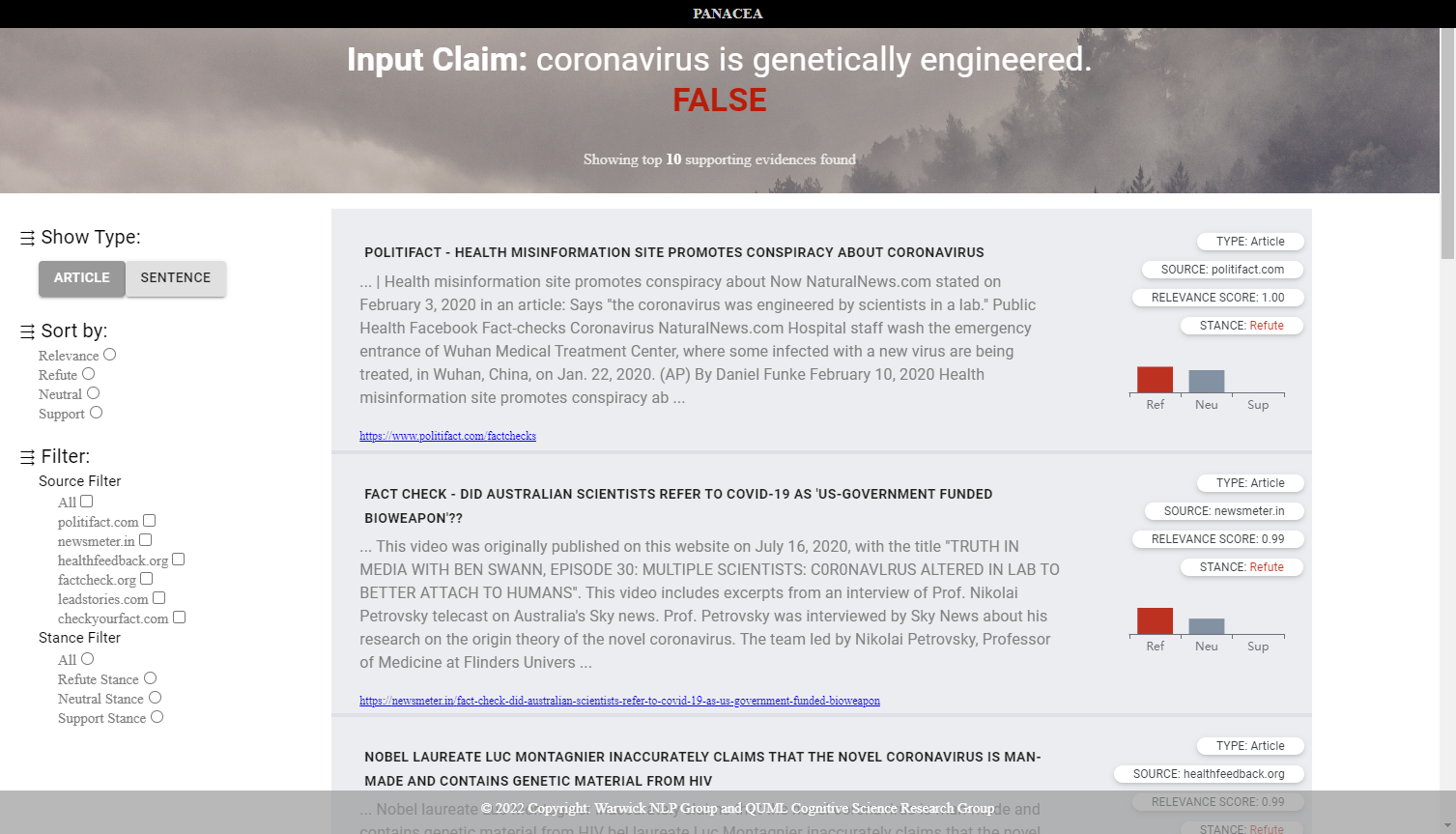}
    \caption{Fact checking result with input claim: \emph{coronavirus is genetically engineered.}}
    \label{fig:fact-check-results}
\end{figure*}

\paragraph{Resource Allocation System}
Users can input natural language claims into our system, and PANACEA provides autocompleted input guesses based on the current input and the claims dataset. %The claims in the dataset that contain current input are shown as in Figure~\ref{fig:autocomplete}. 
Claim autocompletion can help users to input the claim faster and the results included within the claims dataset can be pre-computed for faster retrieval.
%The results for these claims are pre-computed for faster retrieval, 
However, if the user cannot find what they would like to check through the claims dataset, the new claim would be passed to our model for real-time evaluation. Veracity assessment and evidence retrieval are based on our natural language inference model NLI-SAN \cite{nli2022}, which needs GPU resources to run. Therefore we built a queuing system that manages the resources and queues the claims while the GPUs are being used. The results are sent to the user. To avoid duplicate searches, a temporary copy of this result is saved in our database based on the user’s IP address until the user searches for a new claim or the saved period expires.

\begin{figure*}[htb]
    \centering
    \includegraphics[width=0.95\linewidth]{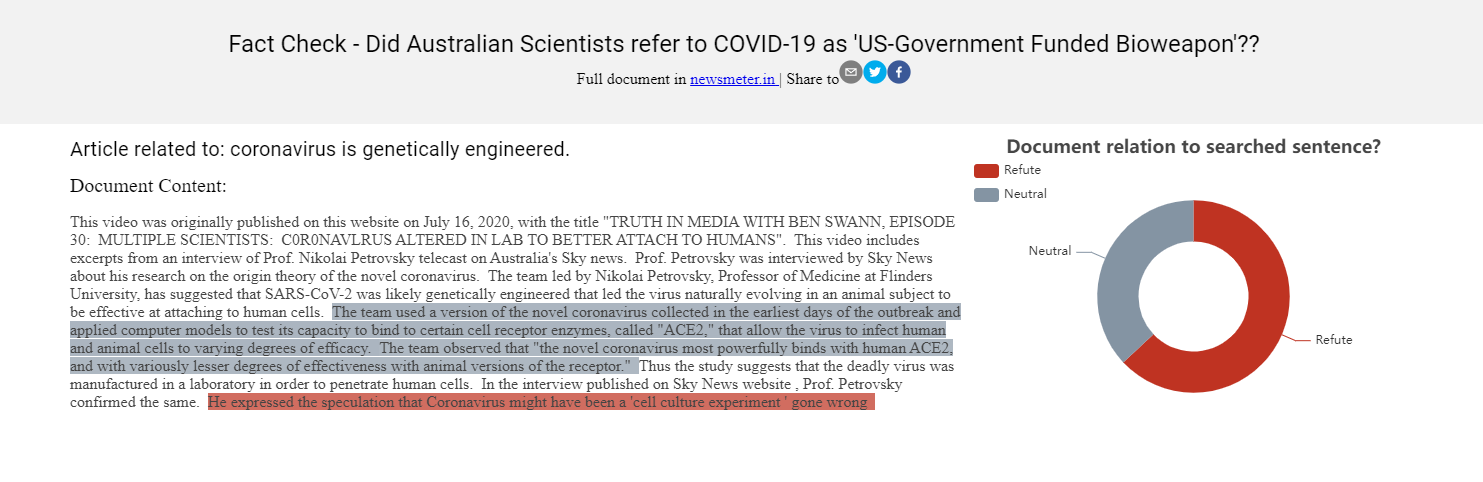}
    \caption{The detail page of user selected supporting document}
    \label{fig:fact-check-details}
\end{figure*}

\paragraph{Veracity Assessment}
PANACEA is supported by NLI-SAN \cite{nli2022}, which incorporates natural language inference results of claim-evidence pairs into a self-attention network. The input claim $c$ is paired with each retrieved relevant evidence $e_i$ to form claim-evidence pairs, where the relevant evidences are the retrieved sentences as described in the following paragraph. 
Each claim-evidence pair $(c, e_i)$ is fed into both a RoBERTa-large\footnote{\label{huggingface} \url{https://huggingface.co/}} model to get a representation $S_i$ and into a RoBERTa-large-MNLI$^{\ref{huggingface}}$ model to get a probability triplet $I_i$ of stance (\emph{contradiction}, \emph{neutrality}, or \emph{entailment}) between the pair.
Next, $S_i$ is mapped to a Key $K$ and a Value $V$, while $I_i$ is mapped onto a Query $Q$. $(Q,K,V)_i$ forms the input of the self-attention layer and the outputs $O_i$ for all the claim-evidence pairs are concatenated together. The output is then passed to a MLP layer to get the veracity assessment result (\emph{True} or \emph{False}) as shown in Figure~\ref{fig:fact-check-results}.

\paragraph{Supporting Evidence Retrieval}
This module includes three parts: document retrieval, sentence retrieval and corresponding meta-data generation. Multi-stage retrieval is applied, retrieving first the top 100 relevant documents with BM25, that then are re-ranked by MonoT5 \cite{monot5} and the top 10 documents are selected. For each of those documents, the top 3 sentences are selected. Both documents and sentences are ranked by their relevance score, which is the cosine similarity between the documents/sentences and the input claim embeddings. Each of those texts are represented through embeddings obtained using Sentence-Transformers with the pre-trained model MiniLM-L12-v2 \cite{minilm2020}. 
%The MiniLM-L12-v2 model has not been fine-tuned because there is a shortage of labelled data for similarity tasks. However, it could be enhanced by future research that involves fine-tuning on sentence pairs with a contrastive learning objective or using labelled datasets.
The corresponding metadata of the supporting documents, including type, source, relevance score, and stance towards the claim are also shown, together with the ranked documents/sentences. Users can also filter or re-rank the result using the metadata. An example of documents retrieved is shown in Figure~\ref{fig:fact-check-results} and the corresponding detailed information visualisation is shown in Figure~\ref{fig:fact-check-details}. On the details page, the whole document text is shown with the top 3 relevant sentences highlighted by their stance towards the input claim. The stance distribution, described in the veracity assessment module is also visualised.

\subsection{Rumour Detection}
Another approach to detecting rumours that has been found to be effective \cite{tdrvnn2018, duck2022} is modelling user comments and propagation networks. Next we describe the relevant rumour detection modules of our system.

\begin{figure*}[!htb]
    \centering
    \includegraphics[width=\linewidth]{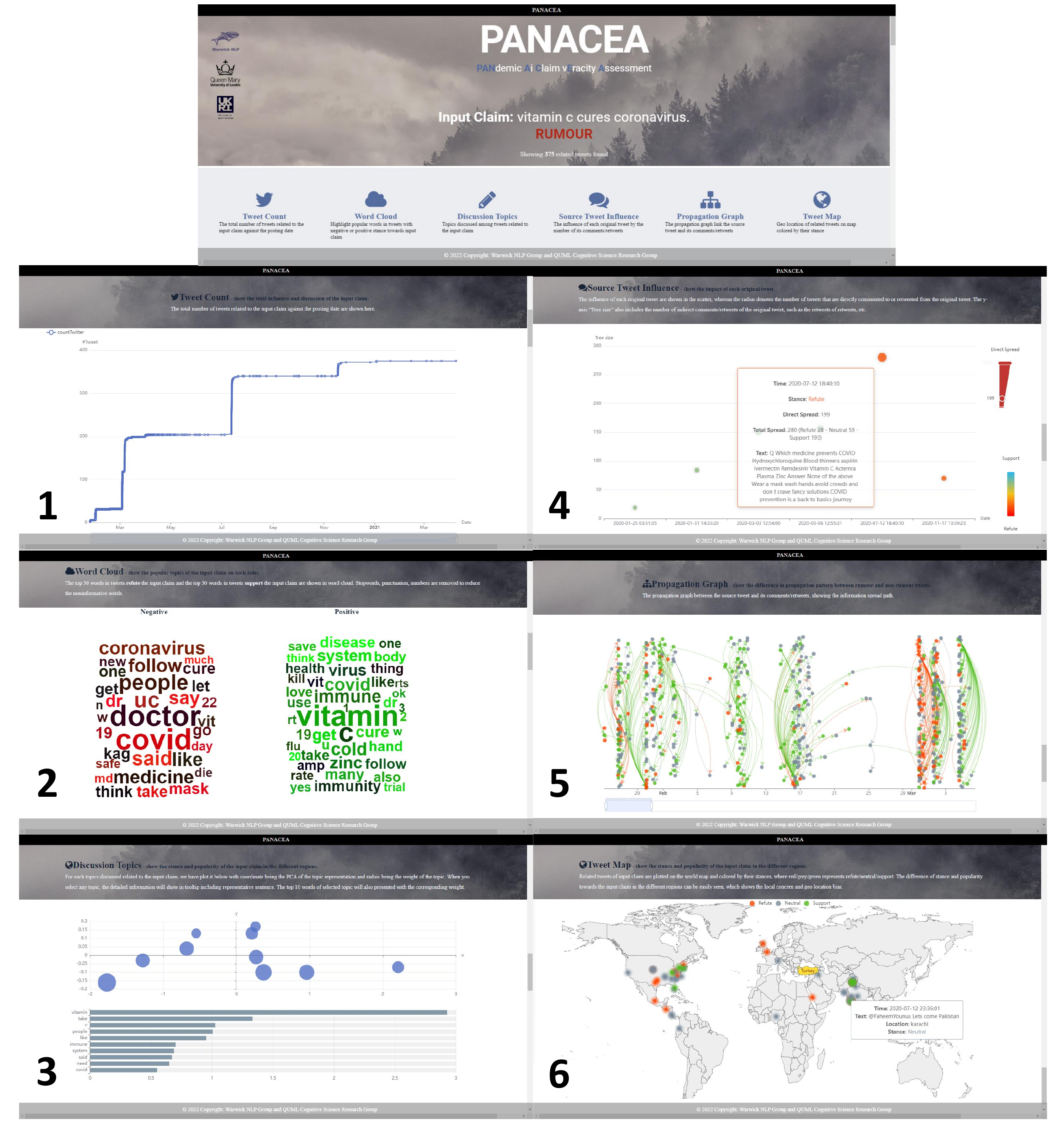}
    \caption{Rumour detection result with input claim: \emph{vitamin c cures coronavirus.}}
    \label{fig:rumour-detection}
\end{figure*}

\paragraph{Claim-related tweets retrieval}
Similar to the fact-checking module, this module includes an autocomplete function for the user's natural language input claim that guesses the input from our claims dataset. The results for existing claims are also pre-computed to retrieve tweets faster. 
For a claim that is not in our claim dataset, we use BM25 to retrieve the related propagation trees from the large Twitter propagation tree database maintained by the active Twitter crawler. 

\paragraph{Rumour Assessment and Data Analysis}
PANACEA adapts a bi-directional graph convolutional networks model (BiGCN) \cite{bigcn2020} to perform rumour detection, which is trained on Twitter16 and fine-tuned on our annotated propagation trees. The reason we chose BiGCN is that it behaves relatively better compared with other models in cross-dataset evaluation \cite{rumourdetection2022}.
For an input claim, the system gives the rumour detection result generated by the weighted average of propagation trees’ rumour assessment label, $\frac{\sum_{i \in T}{n_ir_i}}{\sum_{j \in T}{n_j}}$, where $T$ is the set of retrieved propagation trees.
We generate the sentiment labels of each tweet by VADER\footnote{\url{https://www.nltk.org/api/nltk.sentiment.vader.html}} and stance of tweet towards the input claims by natural language inference \cite{ANLI2020}.

\paragraph{Twitter propagation visualisation}
%\paragraph{Twitter meta-information analysis}
As shown in Figure~\ref{fig:rumour-detection}, PANACEA has six modules, which use the metadata we crawled from the tweet and generated from data analysis to visualise the propagation pattern: %(1) Tweet Count. This module shows the total number of tweets related to the input claim against the posting date, and aims to reflect the total influence and discussion of the input claim. (2) Word Cloud. This module shows the top 50 words in tweets that refute the input claim and the top 50 words in tweets that support the input claim. Stopwords, punctuation, and numbers are removed to reduce the non-informative words. (3) Discussion Topics. This topic build on Latent Dirichlet Allocation (LDA), where each topic are encoded by COVID-Twitter-BERT \footnote{\url{https://huggingface.co/digitalepidemiologylab/covid-twitter-bert-v2}} and the representative tweet is selected by its embedding similarity with the topic. The principal component analysis (PCA) is applied to visualise each topic. Top 10 words and corresponding weights of chosen topic are shown in bar chart. (4) Tweet Spread. This module shows the influence of each original tweet in the scatter plot, whereas the radius denotes the number of tweets that are direct comments or retweets from the original tweet. The y-axis “Total Spread” also includes the number of indirect comments/retweets of the original tweet, such as the retweets of retweets, etc.  (5) Propagation Graph.  This module shows the propagation graph between the source tweet and its comments, showing the information spread path. 5 other claims are randomly chosen from popular claims for users to compare propagation patterns. This module aims to visualise propagation graphs in a straightforward way and let users see the difference between trees of different types. (6) Tweet Map. Related tweets to the input claim are plotted on the world map and coloured by their stances, where \emph{red/yellow/blue} represents \emph{refute/neutral/support}. The difference in stance and popularity towards the input claim in the different regions can be easily seen, which shows the local concern and geo-location bias.
\begin{enumerate}
\itemsep0em 
\item \underline{Tweet Count}, showing the total number of tweets related to the input claim against the posting date, and aiming to reflect the total influence and scale of discussion of the claim. 
\item \underline{Word Cloud}, showing the top 30 words in tweets refuting the input claim and the top 30 words in tweets supporting the input claim. Stopwords, punctuation, and numbers are removed to reduce  non-informative words. 
\item \underline{Discussion Topics}, building on Latent Dirichlet Allocation (LDA), where each topic is encoded by COVID-Twitter-BERT \footnote{\url{https://huggingface.co/digitalepidemiologylab/covid-twitter-bert-v2}} and the representative tweet is selected by its embedding similarity with respect to the topic. Principal component analysis (PCA) is applied to visualise each topic. Top 10 words and corresponding weights of the chosen topic are shown in a bar chart. 
\item \underline{Tweet Spread}, showing the influence of each original tweet in the scatter plot, where the radius denotes the number of tweets that are direct comments or retweets from the original tweet. The y-axis “Total Spread” also includes the number of indirect comments/retweets of the original tweet, such as the retweets of retweets, etc.  
\item \underline{Propagation Graph}, showing the propagation graph between the source tweet and its comments, showing the information spread path. 5 other claims are randomly chosen from popular claims for users to compare propagation patterns. This module aims to visualise propagation graphs in a straightforward way and help users see the difference between trees of different types. 
\item \underline{Tweet Map}. Related tweets to the input claim are plotted on the world map and coloured by their stances, where \emph{red/yellow/blue} represents \emph{refute/neutral/support}. The difference in stance and popularity towards the input claim in the different regions can be easily seen, which shows the local context and geo-location bias.
\end{enumerate}

\section{Evaluation Results}
\begin{figure}[htb]
    \centering
    \includegraphics[width=\linewidth]{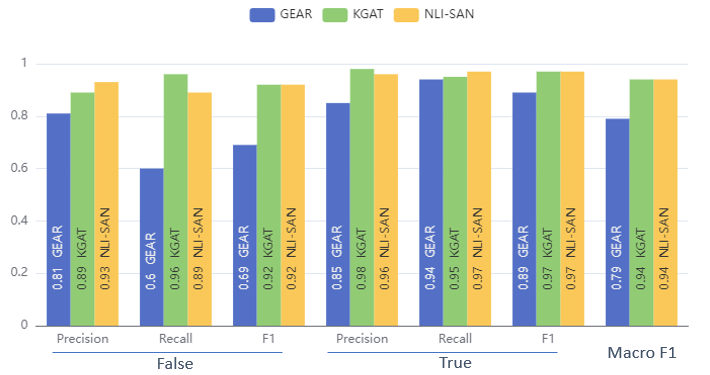}
    \caption{Veracity classification on the PANACEA dataset.}
    \label{fig:nlisan}
\end{figure}

\paragraph{Fact-Checking}
We investigate the performance of our system in document retrieval and veracity assessment in \cite{nli2022}. Table~\ref{tab:retrieval} shows that combining BM25 and MonoT5 is the most effective approach for document retrieval of the selected techniques. In addition, Figure~\ref{fig:nlisan}  shows that NLI-SAN achieves similar performance with KGAT \cite{KGAT2020}, %while being simpler and more efficient 
while having a simpler architecture for the application, and outperforms GEAR \cite{Gear2019}. 

\begin{table}[htb]
\centering
\resizebox{\columnwidth}{!}{
\begin{tabular}{ccccc}
\toprule
\textit{}       & AP@5          & AP@10         & AP@20         & AP@100        \\\midrule
BM25            & 0.54          & 0.56          & 0.58          & 0.62          \\
BM25+MonoBERT   & 0.52          & 0.55          & 0.58          & 0.62          \\
BM25+MonoBERT   & \textbf{0.55} & \textbf{0.58} & \textbf{0.60} & \textbf{0.62} \\
BM25+RM3+MonoT5 & 0.51          & 0.53          & 0.55          & 0.57         \\ \bottomrule
\end{tabular}}
\caption{Document retrieval on the PANACEA dataset.}
\label{tab:retrieval}
\end{table}

\begin{figure}[htb]
    \centering
    \includegraphics[width=\linewidth]{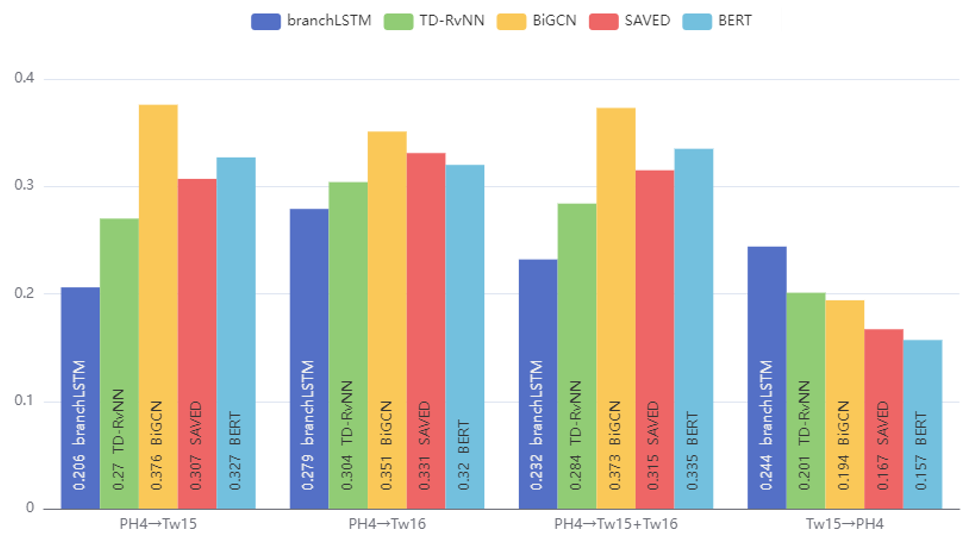}
    \caption{Cross-dataset evaluation of models train and test on different datasets, such as training on PHEME, testing on Twitter15/Twitter16 and vice versa.}
    \label{fig:rumour_result}
\end{figure}

\paragraph{Rumour Detection}
As shown in Figure~\ref{fig:rumour_result}, our comparison \cite{rumourdetection2022} among various models, including branchLSTM \cite{branchLSTM2020}, TD-RvNN \cite{tdrvnn2018}, BiGCN \cite{bigcn2020}, SAVED \cite{saved2021} and BERT \cite{bert2019} for rumour detection evaluated on Twitter15, Twitter16 and PHEME \cite{pheme}, reveals there is no model that always performs the best. Although state-of-the-art models can achieve high accuracy on their training datasets, such performance drops quickly while evaluating on a different dataset \cite{rumourdetection2022}. Due to the limitation of existing models in generalisation, users should interpret this result with caution as the system cannot guarantee output correctness.

\section{Conclusion}
This paper introduces a web-based system on \emph{fact-checking} and \emph{rumour detection} based on novel natural language processing models for COVID-19 misinformation detection. Going forward, we will keep updating the data and explore other methods for misinformation identification to improve the current system and introduce more functions to the system as part of our continuing efforts to support %our project goal of helping unprofessional users 
the general public to identify misinformation.

\section*{Acknowledgements}
This work was supported by the UK Engineering and Physical Sciences Research Council (grant no. EP/V048597/1). YH and ML are each supported by a Turing AI Fellowship funded by the UK Research and Innovation (grant no. 	EP/V020579/1, EP/V030302/1).

\bibliography{anthology,custom}
\bibliographystyle{acl_natbib}

\appendix

%\section{Example Appendix}
%\label{sec:appendix}

%This is an appendix.

\end{document}